%% file: main.tex
\documentclass{article}

\usepackage{epsfig}
\usepackage{graphicx}
\usepackage{amsmath}
\usepackage{amssymb}
\usepackage{color}
\usepackage{algorithmic}
\usepackage{algorithm}
\usepackage{comment}
\usepackage{subcaption}
\usepackage[breaklinks=true,colorlinks,bookmarks=false,citecolor={cyan}]{hyperref}
\usepackage{authblk}
\usepackage[margin=1.2in]{geometry}
\usepackage{microtype}
\usepackage[square, numbers, comma, sort&compress]{natbib}
\usepackage{hyperref}

\newcommand{\emc}[1]{{\textbf{\textit{\color[rgb]{0,.3,.6}#1}}}}

\DeclareMathOperator{\vecc}{vec} 
\DeclareMathOperator{\unvecc}{unvec} 
\DeclareMathOperator{\diag}{diag}

\begin{document}

\title{Data Augmentation via Structured Adversarial Perturbations}

\author{
\vspace{0.2in}
Calvin Luo\thanks{Work done as part of the Google Research residency program.}$^{*}$ \quad Hossein Mobahi \quad Samy Bengio\\

{\color{magenta}\texttt{\{calvinluo,hmobahi,bengio\}@google.com}}\\
\vspace{0.2in}
Google Research\\
Mountain View, CA, USA\\
}
\date{}

\maketitle
\begin{abstract}
\input{sections/abstract}
\end{abstract}

\input{sections/intro}
\input{sections/related}

\input{sections/method}
\input{sections/experiments}
\input{sections/discussion}

\section*{Acknowledgements}
We would like to thank Andrew Ilyas and Nicholas Carlini for their comments.

\bibliographystyle{apalike}
\bibliography{references}

\end{document}

%% file: sections/abstract.tex
Data augmentation is a major component of many machine learning methods with state-of-the-art performance. Common augmentation strategies work by drawing random samples from a space of transformations. Unfortunately, such sampling approaches are limited in expressivity, as they are unable to scale to rich transformations that depend on numerous parameters due to the curse of dimensionality. Adversarial examples can be considered as an alternative scheme for data augmentation. By being trained on the most difficult modifications of the inputs, the resulting models are then hopefully able to handle other, presumably easier, modifications as well. The advantage of adversarial augmentation is that it replaces sampling with the use of a single, calculated perturbation that maximally increases the loss. The downside, however, is that these raw adversarial perturbations appear rather unstructured; applying them often does not produce a natural transformation, contrary to a desirable data augmentation technique. To address this, we propose a method to generate adversarial examples that maintain some desired natural structure. We first construct a subspace that only contains perturbations with the desired structure. We then project the raw adversarial gradient onto this space to select a structured transformation that would maximally increase the loss when applied. We demonstrate this approach through two types of image transformations: photometric and geometric. Furthermore, we show that training on such structured adversarial images improves generalization.

%% file: sections/intro.tex
\section{Introduction}
\label{sec:intro}

In many machine learning applications, data augmentation is integral for achieving  state-of-the-art performance \cite{touvron2020fixing, xie2020self, chen2020simple}. Most augmentation strategies used in common practice consider a parameterized space of transformations. Then, to generate an augmented image, random \emc{samples} from the parameter space are drawn, and the resulting associated transformation is applied to the input data. Unfortunately, for many interesting transformations, the parameter space is large, and sampling can become inefficient or even intractable due to the \emc{curse of dimensionality}. As a result, current popular data augmentation strategies in practical use today are defined by only a few parameters, and are therefore limited in the richness of their transformations.

Adversarial examples can be interpreted as a form of data augmentation.  In its original formulation, an adversarial example is a datum that has been intentionally perturbed in such a way as to fool the classifier \cite{goodfellow2014explaining, szegedy2013intriguing, kurakin2016adversarial, dong2018boosting}.  By being trained on the most difficult modifications on the inputs, the resulting models are then hopefully able to handle other, presumably easier, modifications as well.

The ultimate purpose of data augmentation is to provide the model with more examples of natural data in a cheap way.  To accomplish this, we desire highly \emc{structured} transformations on our existing inputs.  In image classification, for example, we may use translations, rotations, or scaling transforms to approximate real movements of the objects depicted in the image.  To this end, vanilla adversarial examples fall short in terms of an augmentation procedure; the adversarial perturbation is often unstructured, as its sole goal is to break the classifier rather than to produce a naturally transformed version of the input.  Furthermore, training on adversarial examples has been observed in practice to either negligibly improve or even lower test accuracy \cite{santurkar2020breeds}.  We conjecture that this happens because learning to handle adversarial examples is an excessive endeavor; it wastes a portion of the model's capacity to learn difficult transformations that may not occur in natural data.

\begin{figure}
\centering
\begin{tabular}{c c c c}
\includegraphics[width=.22\linewidth]{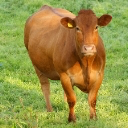} & \includegraphics[width=.22\linewidth]{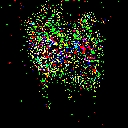} & \includegraphics[width=.22\linewidth]{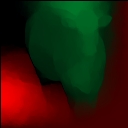} & \includegraphics[width=.22\linewidth]{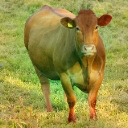} \\
{\bf (a)} & {\bf (b)} & {\bf (c)} & {\bf (d)}
\end{tabular}
\caption{Visualizing the recolorization transformation for a toy cow image. {\bf (a)}~ The original image. {\bf (b)}~ The raw adversarial perturbation $\nabla_{\boldsymbol{z}} L$ on a ResNet model trained on ImageNet. {\bf (c)}~ Generated transformation by projecting the raw perturbation onto the recolorization space (Section~ \ref{sec:recolor}). {\bf (d)}~ Application of the generated transformation to the image. The ResNet model is only 61.16\% confident that the image contains a cow, compared to 87.29\% for the original image.}
\label{fig:cow}
\end{figure}

In this work, we propose a  method to generate adversarial examples that maintain some desired natural structure. We first construct a subspace that only contains perturbations with the desired structure. Then, we project the raw adversarial gradient onto this space to select structured transformations that would maximally increase the loss when applied (see Figure~\ref{fig:cow}).

We illustrate this approach through two types of image transformations: photometric and geometric. Additionally, we demonstrate that training on such structured adversarial images improves generalization.

%% file: sections/related.tex
\section{Related Works}

Recent work has explored learning augmentation strategies that best fit the dataset at hand, rather than using manually designed ones.  Cubuk et al. \cite{cubuk2019autoaugment} propose AutoAugment, a reinforcement learning method that automatically learns an improved data augmentation policy.  Because searching for an optimal policy is computationally expensive, followup work by Cubuk et al. \cite{cubuk2019randaugment} fix the policy to simply apply a series of uniformly-selected transformations.  This not only reduces computational complexity, but also improves performance under certain datasets. Zhang et al. \cite{zhang2019adversarial} propose using a computationally-affordable AutoAugment policy that adapts to a network as it is being trained, creating consistently hard images for the model throughout the optimization process.
Ratner et al. \cite{ratner2017learning} frames data augmentation as sequence modeling over user-specified transformation functions, and learns to generate novel transformation combinations through a generative adversarial network.  However, all of these proposed learning-based augmentation methods rely on sampling-based augmentations, and may not scale to richer transformations.

The use of auxiliary networks for data augmentation has also been proposed.  Rather than relying on combining existing augmentation strategies, Zhao et al. \cite{zhao2019data} learn a transformation network from a single labelled example and many unlabelled examples in a semi-supervised fashion.  This network, modeled as a U-Net \cite{ronneberger2015u}, is then able to synthesize novel training images.  Poursaeed et al. \cite{poursaeed2018generative} train U-Nets to produce both universal and image-dependent perturbations to perform targeted and untargeted attacks for a given image.  Jena et al. \cite{jena2020ma3} propose utilizing a Spatial Transformer Network \cite{jaderberg2015spatial} to produce adversarial transformation parameters that warp the image into an example that is difficult for the model to classify.  Bin et al. \cite{bin2020adversarial} also use a Spatial Transformer Network as a generator to augment an image of a human with additional segmented body parts in an adversarial way, and a discriminator to learn the correct pose.  Despite their ability to generate novel images, the added cost in model size and training time from these additional networks makes them unattractive for use as data augmentation techniques.

Recent work has also explored training models on manipulations of the embedded feature space, which can be interpreted as implicitly processing semantically transformed data samples.  Many of these approaches rely on a Generative Adversarial Network \cite{goodfellow2014generative} or a generative adversarial learning scheme in order to ensure that the manipulated features remain valid representations of a natural image\cite{li2020adversarial, volpi2018adversarial}.  Not only is the presence of these auxiliary networks undesirable, but the manipulations in the feature space are generated via sampling noise vectors rather than some truly structured adversarial way with respect to the classifier.  Liu et. al \cite{liu2018feature} use an encoder-decoder architecture to learn to generate valid variations in perspective for data augmentation, but is limited to domains where pose information is available.  Wang et. al \cite{wang2019implicit} estimate the covariance matrix of deep features for each class to determine semantically meaningful intra-class dimensions that can be modified.  However, this technique also relies on sampling to generate an augmentation, and the online estimation of the covariance matrix is an expensive iterative procedure.

Adversarial training, where the classifier is trained with adversarial examples, has been shown to improve robustness to such attacks \cite{goodfellow2014explaining, madry2017towards}. Song et al. \cite{song2018improving} formulate the adversarial training learning procedure as a domain adaptation problem, and improve the resulting learned model's generalization ability across different adversarial attacks.  Xie et al. \cite{xie2019adversarial} also perform adversarial training, but learn separate batch normalization \cite{ioffe2015batch} parameters for adversarial examples due to their different underlying distribution.

To generate adversarial examples, iterative methods have been suggested.  Carlini et al. \cite{carlini2017towards} optimize for the smallest image perturbation that would result in a misclassification.  Manifool \cite{kanbak2018geometric} generates a misclassified example for a given image and transformation function by iteratively producing augmentations that move towards the decision boundary while remaining on the transformation manifold. The transformation parameters of these augmentations are created by projecting the image gradient onto the tangent space of the manifold.  Xiao et al. \cite{xiao2018spatially} learn a flow field that can spatially transform an image into an adversarial example.  These methods are expensive due to the iterative nature, and thus infeasible for use as a data augmentation scheme.

Recent works have tried to modify adversarial examples to be structured.  Qiu et al. \cite{qiu2019semanticadv} generate semantically meaningful synthetic images by feeding them through an attribute-conditioned image generator with a perturbed attribute vector.  The adversarial image is then created by interpolating the synthetic and original images in order to make it appear more natural.  Bhattad et al. \cite{bhattad2019big} leverage CNN-based texture transfer methods and colorization models to generate adversarial attacks.  They also impose regularizers in their objective to keep the perturbation semantically meaningful and the attack image realistic.  Song et al. \cite{song2018constructing} propose using a generative model to synthesize entirely novel adversarial images instead of perturbing an existing one.  They accomplish this by searching over the latent space to find images that will be misclassified.  Joshi et al. \cite{joshi2019semantic} optimize an adversarial loss over parameters to feed into a parametric conditional generative model.  However, the addition of these auxiliary networks makes them inherently inefficient as an augmentation procedure.

Other structured perturbations are generated through random procedures rather than optimization.  Hosseini et al. \cite{hosseini2018semantic} attempt to find an adversarial image by randomly changing the color and hue of an image while preserving shapes until they arrive at an adversarial image.  Zhao et al. \cite{zhao2017generating} propose generating natural adversarial images by randomly sampling from the neighborhood of a latent embedding and then decoding it.  However, these sampling procedures gives us little control over and limited interpretability about the resulting perturbed image.  Furthermore, the sampling process may still be too computationally expensive for a data augmentation technique.

%% file: sections/method.tex
\section{Setup}

Let us define a loss function $L(\boldsymbol{w}\,;\,\boldsymbol{z},\boldsymbol{l})$ that takes as its input model parameters $\boldsymbol{w}$, input image $\boldsymbol{z}$, and target label $\boldsymbol{l}$. Suppose components of $\boldsymbol{z}$ (i.e. pixels) are sampled from a continuous image flow $Z(x,y,t)$ at a specific $t$,
\begin{equation}
\boldsymbol{z} \triangleq [Z(x_1,y_1,t),\dots,Z(x_n,y_m,t)] \,.
\end{equation}
Then, the loss can be equivalently expressed as,
\begin{equation}
L(Z(x_1,y_1,t),\dots,Z(x_n,y_m,t)) \,,
\end{equation}
where, with abuse of notation, we have dropped arguments $\boldsymbol{w}$ and $\boldsymbol{l}$ for brevity. To generate an adversarial example, we seek a perturbation in the image that maximally increases the loss. The rate of change of the loss due to the perturbation can be expressed as,
\begin{equation}
\label{eq:loss_pert}
\frac{\partial}{\partial t} L = \langle \nabla_{\boldsymbol{z}} L(\boldsymbol{z}) \,,\, [\dot{Z}(x_1,y_1,t),\dots,\dot{Z}(x_n,y_m,t)] \rangle\,,
\end{equation}
where, $\dot{Z}(x,y,t)\triangleq \frac{\partial}{\partial t}Z(x,y,t)$. Maximization of the above quantity amounts to maximizing the dot product of the gradient of the loss and image perturbation  $\dot{\boldsymbol{z}}$. In a typical adversarial example scenario, $\dot{\boldsymbol{z}}$ is restricted by some $p$-norm of a vector such that $\|\dot{\boldsymbol{z}}\|_p \leq \epsilon$ for some choice of $\epsilon>0$. In the following sections we demonstrate ways to incorporate geometric and photometric structures to construct $\dot{z}$.

\section{Geometric Transform}

We leverage the \emc{brightness constancy} assumption, which presumes the brightness value of a small region remains constant despite its change in location \cite{Horn81},
\begin{equation}
\frac{d}{d t} Z(x(t),y(t),t) = 0\,,
\end{equation}
which using chain rule can be expressed as,
\begin{equation}
\label{eq:brightness_constancy}
Z_x(x(t),y(t),t)\dot{x}(t) \,+\, Z_y(x(t),y(t),t)\dot{y}(t) \,+\, \dot{Z}(x(t),y(t),t) \,=\, 0\,.
\end{equation}
Applying this to (\ref{eq:loss_pert}) yields,
\begin{eqnarray}
\frac{\partial}{\partial t} L = - \langle \, \nabla_{\boldsymbol{z}} L(\boldsymbol{z}) \,,\,[ & & {Z}_x(x_1,y_1,t)\dot{x}(x_1,y_1,t)  +{Z}_y(x_1,y_1,t)\dot{y}(x_1,y_1,t) \nonumber \\
&,& \nonumber \\
&\vdots& \nonumber \\
&,& Z_x(x_n,y_m,t) \dot{x}(x_n,y_m,t) + Z_y(x_n,y_m,t) \dot{y}(x_n,y_m,t) ] \,\rangle\,. \nonumber
\end{eqnarray}
We are interested in the evaluation of perturbation at the reference point $t=0$. For brevity, denote $Z_x(x_i,y_j,t=0)$ by $Z_{x,i,j}$ and  $\dot{x}((x_i,y_j),t=0)$ by $\dot{x}_{i,j}$. Also, denote each component of $\nabla_{\boldsymbol{x}}L(\boldsymbol{x})$ by $\ell_{i,j}$. We can now express the above identity in vector/matrix form.
\begin{equation}
\frac{\partial}{\partial t} L = -\begin{pmatrix}  \boldsymbol{d}_x\\
\boldsymbol{d}_y
\end{pmatrix}^T \begin{pmatrix}
\dot{\boldsymbol{x}} \\
\dot{\boldsymbol{y}}
\end{pmatrix}
\end{equation}
where,
\begin{eqnarray}
N &\triangleq& n \, m \\
{\boldsymbol{d}_x}_{\color{red} N \times 1} &\triangleq& \begin{pmatrix} 
\ell_{1,1} Z_{x,1,1} & \dots & \ell_{m,n} Z_{x,m,n} 
\end{pmatrix} \\
{\boldsymbol{d}_y}_{\color{red} N \times 1} &\triangleq& \begin{pmatrix} 
\ell_{1,1} Z_{y,1,1} & \dots & \ell_{m,n} Z_{x,m,n} 
\end{pmatrix} \\
\dot{\boldsymbol{x}}_{\color{red} N \times 1} &\triangleq& \begin{pmatrix} 
\dot{x}_{1,1} & \dots & \dot{x}_{m,n} 
\end{pmatrix} \\
\dot{\boldsymbol{y}}_{\color{red} N \times 1} &\triangleq& \begin{pmatrix} 
\dot{y}_{1,1} & \dots & \dot{y}_{m,n}
\end{pmatrix}
\end{eqnarray}

Our goal is to find a motion field characterized by $\dot{\boldsymbol{x}}$ and $\dot{\boldsymbol{y}}$ that maximizes $\frac{\partial}{\partial t} L$. However, as the problem is unbounded in this form, we have to restrict the norms $\| \dot{\boldsymbol{x}} \|$ and  $\| \dot{\boldsymbol{y}} \|$. In addition, we would like the motion field to have some coherency and smoothness. Hence, we cast the problem as the following optimization,
\begin{equation}
(\dot{\boldsymbol{x}}^* , \dot{\boldsymbol{y}}^*) \,=\,  \arg\min_{(\dot{\boldsymbol{x}} , \dot{\boldsymbol{y}})} \begin{pmatrix}  \boldsymbol{d}_x \\
\boldsymbol{d}_y
\end{pmatrix}^T \begin{pmatrix}
\dot{\boldsymbol{x}} \\
\dot{\boldsymbol{y}}
\end{pmatrix} \,+\, \frac{1}{2} \alpha(\| \dot{\boldsymbol{x}} \|^2+\| \dot{\boldsymbol{y}} \|^2) \,+\, \frac{1}{2} \beta(\| \boldsymbol{D}_x \dot{\boldsymbol{x}} \|^2 + \| \boldsymbol{D}_y \dot{\boldsymbol{x}} \|^2 ) \,+\, \frac{1}{2} \beta( \| \boldsymbol{D}_x \dot{\boldsymbol{y}} \|^2+\| \boldsymbol{D}_y \dot{\boldsymbol{y}} \|^2) \,, \nonumber
\end{equation}
where $\boldsymbol{D}_x$ and $\boldsymbol{D}_y$ are differential operators along $x$ and $y$ directions, represented in matrix form, and $\alpha$ and $\beta$ are scalars. For brevity, let $\boldsymbol{P} \triangleq \boldsymbol{D}_x^T \boldsymbol{D}_x + \boldsymbol{D}_y^T \boldsymbol{D}_y $. We obtain,
\begin{equation}
(\dot{\boldsymbol{x}}^* , \dot{\boldsymbol{y}}^*) \,=\,  \arg\min_{(\dot{\boldsymbol{x}} , \dot{\boldsymbol{y}})} \begin{pmatrix}  \boldsymbol{d}_x\\
\boldsymbol{d}_y
\end{pmatrix}^T \begin{pmatrix}
\dot{\boldsymbol{x}} \\
\dot{\boldsymbol{y}}
\end{pmatrix} \,+\, \frac{1}{2}\alpha( \dot{\boldsymbol{x}}^T \boldsymbol{I} \dot{\boldsymbol{x}} + \dot{\boldsymbol{y}}^T \boldsymbol{I} \dot{\boldsymbol{y}}) \,+\, \frac{1}{2} \beta( \dot{\boldsymbol{x}}^T \boldsymbol{P} \dot{\boldsymbol{x}} + \dot{\boldsymbol{y}}^T \boldsymbol{P} \dot{\boldsymbol{y}}) \,.
\end{equation}
Letting $\gamma \triangleq \frac{\beta}{\alpha}$, the above can be equivalently written as,
\begin{equation}
(\dot{\boldsymbol{x}}^* , \dot{\boldsymbol{y}}^*) \,=\,  \arg\min_{(\dot{\boldsymbol{x}} , \dot{\boldsymbol{y}})}  \begin{pmatrix}  \boldsymbol{d}_x\\
\boldsymbol{d}_y
\end{pmatrix}^T \begin{pmatrix}
\dot{\boldsymbol{x}} \\
\dot{\boldsymbol{y}}
\end{pmatrix}\\
\,+\, \frac{1}{2}\alpha\big( ( \dot{\boldsymbol{x}}^T \boldsymbol{I} \dot{\boldsymbol{x}} + \dot{\boldsymbol{y}}^T \boldsymbol{I} \dot{\boldsymbol{y}}) \,+\, \gamma( \dot{\boldsymbol{x}}^T \boldsymbol{P} \dot{\boldsymbol{x}} + \dot{\boldsymbol{y}}^T \boldsymbol{P} \dot{\boldsymbol{y}}) \big) \,.
\end{equation}
 
\begin{figure}
\centering
\begin{tabular}{c c c c}
\includegraphics[width=.22\linewidth]{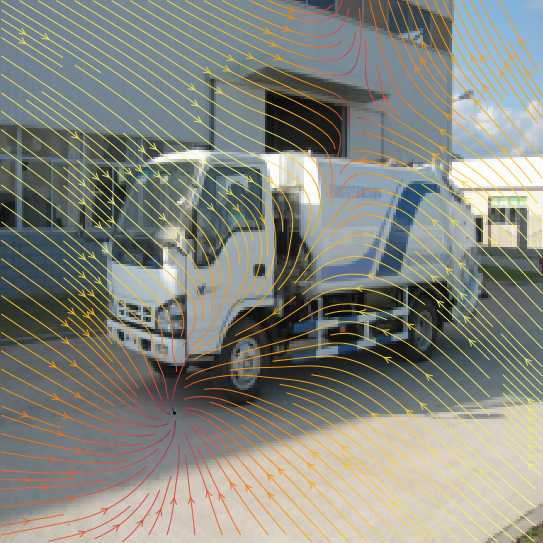} & \includegraphics[width=.22\linewidth]{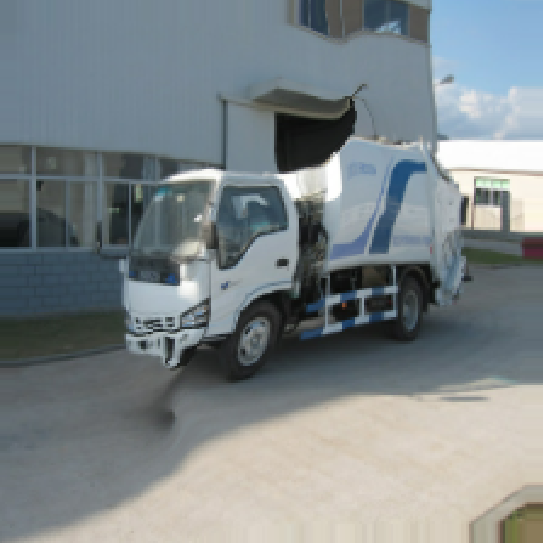} & \includegraphics[width=.22\linewidth]{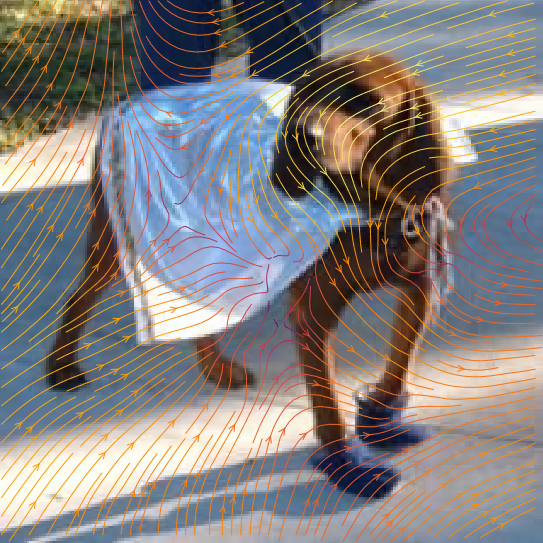} & \includegraphics[width=.22\linewidth]{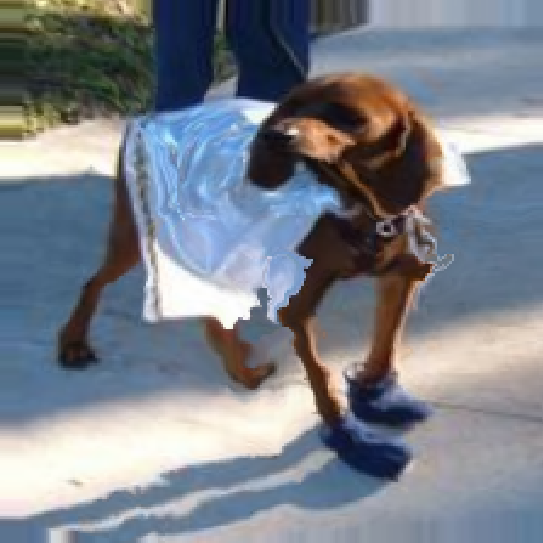} \\
{\bf (a)} & {\bf (b)} & {\bf (c)} & {\bf (d)}
\end{tabular}
\caption{Visualizing the effect of adversarial flow on ImageNet examples, using a trained ResNet model.  {\bf(a).}~A garbage truck image correctly classified with a confidence of 60.95\%, and overlaid with its computed flow field. {\bf(b).}~ After applying the motion flow field, the model predicts the image to be of a minibus, with 33.90\% confidence; the flow warped the back of the truck to be raised - a trait common in minibuses. {\bf(c).}~An image of a redbone (a dog breed) correctly classified with a confidence of 44.95\%. {\bf (d).}~ The warped dog is classified as a boxer (another breed of dog) with 19.34\% confidence.  The flow appears to warp the white cloth onto the dog's leg; redbones are normally entirely red, whereas boxers are commonly red with patches of white fur.}
\label{fig:animals}
\end{figure}

This is a convex objective function in $(\dot{\boldsymbol{x}} , \dot{\boldsymbol{y}})$ and its minimizer can be obtained by zero crossing its gradient,
\begin{equation}
\boldsymbol{d}_x \,+\, \alpha ( \boldsymbol{I} + \gamma \boldsymbol{P}) \dot{\boldsymbol{x}} = \boldsymbol{0} \quad,\quad \boldsymbol{d}_y \,+\, \alpha (\boldsymbol{I} + \gamma \boldsymbol{P}) \dot{\boldsymbol{y}} = \boldsymbol{0} \,.
\end{equation}
Therefore,
\begin{equation}
\dot{\boldsymbol{x}} = - \frac{1}{\alpha} (\boldsymbol{I} + \gamma \boldsymbol{P})^{-1} \boldsymbol{d}_x \quad,\quad
\dot{\boldsymbol{y}} = - \frac{1}{\alpha} (\boldsymbol{I} + \gamma \boldsymbol{P})^{-1} \boldsymbol{d}_y \,.
\end{equation}
Note that the matrix $(\boldsymbol{I} + \gamma \boldsymbol{P})^{-1}$ solely depends on regularizer, and does not depend on data (data being $\boldsymbol{d}_x$ and $\boldsymbol{d}_y$). Thus, the matrix $(\boldsymbol{I} + \gamma \boldsymbol{P})^{-1}$ can be computed \emc{offline} given the image dimensions of the input and the trade off parameter $\gamma$. Once the motion field ($\dot{\boldsymbol{x}},\dot{\boldsymbol{y}})$ is computed, it can be used to warp $I$ into a new image. The resulted algorithm is shown in Algorithm~\ref{alg:geo}, in which $\odot$ denotes element-wise product, and illustrative examples are provided in Figure~\ref{fig:animals}.

\begin{algorithm}
\caption{: \texttt{geometric\_augmentation}}
\label{alg:geo}
\begin{algorithmic}[1]
\STATE{{\bf input:} mini-batch image/label pairs  $\cup_k \{(\boldsymbol{Z}_k,\boldsymbol{l}_k)\}$, weight vector $\boldsymbol{w}_0$, regularization parameters $\alpha>0$ and $\gamma>0$.}
\STATE{$\boldsymbol{A}_{\color{red} N \times N} \leftarrow (\boldsymbol{I}_{\color{red} N\times N} + \gamma \boldsymbol{P}_{\color{red} N\times N})^{-1}$}
\STATE{${\boldsymbol{z}_k}_{\color{red} N \times 1} \leftarrow \vecc({\boldsymbol{Z}_k}_{\color{red} m \times n})$}
\STATE{$\boldsymbol{g}_{\color{red} N\times 1}(\boldsymbol{z}, \boldsymbol{l}) \triangleq \nabla_{\boldsymbol{z}} L(\boldsymbol{w}_0 \,;\, \boldsymbol{z},\boldsymbol{l})$}
\FOR{$k=1$ \TO $K$}
\STATE{${\boldsymbol{g}^\dag}_{\color{red} N \times 1} \leftarrow \boldsymbol{g}\big(\vecc({\boldsymbol{Z}_k}_{\color{red} m \times n})_{\color{red} N \times 1}, \boldsymbol{l}_k\big)$}
\STATE{$\boldsymbol{u}_{\color{red} N \times 1} \leftarrow \vecc(\frac{\partial}{\partial x}{\boldsymbol{Z}_k})$}
\STATE{$\boldsymbol{v}_{\color{red} N \times 1} \leftarrow \vecc(\frac{\partial}{\partial y}{\boldsymbol{Z}_k})$}
\STATE{${\boldsymbol{d}_x}_{\color{red} N \times 1} \leftarrow  \boldsymbol{u} \odot \boldsymbol{g}^\dag$}
\STATE{${\boldsymbol{d}_y}_{\color{red} N \times 1} \leftarrow  \boldsymbol{v} \odot \boldsymbol{g}^\dag$}
\STATE{$\dot{\boldsymbol{x}}_{\color{red} N \times 1} \leftarrow  - \frac{1}{\alpha} \boldsymbol{A} \boldsymbol{d}_x$}
\STATE{$\dot{\boldsymbol{y}}_{\color{red} N \times 1} \leftarrow  - \frac{1}{\alpha} \boldsymbol{A} \boldsymbol{d}_y$}
\STATE{$\dot{\boldsymbol{X}}_{\color{red} m \times n} \leftarrow \unvecc(\dot{\boldsymbol{x}})$}
\STATE{$\dot{\boldsymbol{Y}}_{\color{red} m \times n} \leftarrow \unvecc(\dot{\boldsymbol{y}})$}
\STATE{${\boldsymbol{Z}_k^*}_{\color{red} m \times n} \leftarrow $\texttt{img\_warp}$(\boldsymbol{Z}_k , \dot{\boldsymbol{X}} , \dot{\boldsymbol{Y}})$ }
\ENDFOR
\RETURN{$(\boldsymbol{Z}_1^*, \dots, \boldsymbol{Z}_K^*)$}
\end{algorithmic}
\end{algorithm}

\section{Photometric Transform}
\label{sec:recolor}
In a photometric transform we would like to preserve contours, object boundaries, edges, etc., while changing the colors in the image. In this section we present one possible formulation for such a transform, with emphasis on its simplicity. Let us define the \emc{edginess} of a pixel location $(x, y)$ as a function $E(x,y,t)$ that is high when the pixel is dissimilar from its neighbors, and low when it is similar to its neighbors. Denoting the color channels\footnote{In this work, all experiments are performed in RGB space.} by A, B, and C, we can penalize the total squared rate of change of $E$,
\begin{equation}
P(t) \triangleq \int_{\mathcal{X}} \int_{\mathcal{Y}} \big(\frac{\partial}{\partial t} E(x,y,t) \big)^2 \, d x \, d y \,,
\end{equation}
We penalize changes in areas with no edges; we encourage low-gradient regions to remain as such after the transformation.  We do this by weighting the change with the inverse of its edginess value. The resulted penalty function has the form,
\begin{equation}
P(t) \triangleq \int_{\mathcal{X}} \int_{\mathcal{Y}} \big( \frac{1}{E(x,y,0)}\frac{\partial}{\partial t} E(x,y,t) \big)^2 \, d x \, d y \,.
\end{equation}
Applying chain rule, this can be expressed as,
\begin{eqnarray}
P(t) = \int_{\mathcal{X}} \int_{\mathcal{Y}} \Big( \frac{1}{E(x,y,0)} \big( & & {Z_A}_x(x,y,t) {\dot{{Z}_A}}_x \,+\, {Z_A}_y(x,y,t) {\dot{{Z}_A}}_y \\
&+& {Z_B}_x(x,y,t) {\dot{{Z}_B}}_x \,+\, {Z_B}_y(x,y,t) {\dot{{Z}_B}}_y \\
&+& {Z_C}_x(x,y,t) {\dot{{Z}_C}}_x \,+\, {Z_C}_y(x,y,t) {\dot{{Z}_C}}_y \\
\big) \Big)^2 \, d x \, d y \,. & &
\end{eqnarray}
We evaluate the above penalty at $t=0$ and as previous section, drop explicit reference to $t$ in our notation,
\begin{eqnarray}
P = \int_{\mathcal{X}} \int_{\mathcal{Y}} \Big( \frac{1}{E(x,y)} \big( & & {Z_A}_x(x,y) {\dot{{Z}_A}}_x \,+\, {Z_A}_y(x,y) {\dot{{Z}_A}}_y \\
&+& {Z_B}_x(x,y) {\dot{{Z}_B}}_x \,+\, {Z_B}_y(x,y) {\dot{{Z}_B}}_y \\
&+& {Z_C}_x(x,y) {\dot{{Z}_C}}_x \,+\, {Z_C}_y(x,y) {\dot{{Z}_C}}_y \\
\big) \Big)^2 \, d x \, d y \,. & &
\end{eqnarray}
This integral can be approximated by evaluating the integrand over an \emc{evenly spaced} grid, where spacing is of size $\delta_x$ and $\delta_y$, and coordinates are specified by $(x_i,y_j)$.
\begin{equation}
P \approx \hat{P} \triangleq \delta_x \delta_y \|  \boldsymbol{M}_A \dot{\boldsymbol{z}}_A + \boldsymbol{M}_B \dot{\boldsymbol{z}}_B  + \boldsymbol{M}_C \dot{\boldsymbol{z}}_C \|^2 \,,
\end{equation}
where,
\begin{eqnarray}
{\boldsymbol{M}_{\alpha}}_{\color{red} N \times N}  &\triangleq& \diag({\boldsymbol{z}_\alpha}_x \div \boldsymbol{e}) \boldsymbol{D}_x  + \diag({\boldsymbol{z}_\alpha}_y \div \boldsymbol{e}) \boldsymbol{D}_y \nonumber \\
\boldsymbol{e}_{\color{red} N \times 1} &\triangleq& \|\, [\, {\boldsymbol{z}_A}_x \,|\, {\boldsymbol{z}_A}_y \,|\, {\boldsymbol{z}_B}_x \,|\, {\boldsymbol{z}_B}_y \,|\, {\boldsymbol{z}_C}_x \,|\, {\boldsymbol{z}_C}_y \,]_{\color{red} N \times 6}\,\| \,, \nonumber
\end{eqnarray}
and $\div$ denotes element-wise division\footnote{Notice that components of $\boldsymbol{e}$ are non-negative. In practice, to avoid division by zero, we add a small constant $\epsilon>0$ to all entries of $\boldsymbol{e}$.} of the vectors, and the $\|\,.\,\|$ is applied row-wise.

Let $\boldsymbol{z} \triangleq [\boldsymbol{z}_A^T \,|\, \boldsymbol{z}_B^T \,|\, \boldsymbol{z}_C^T ]^T$. The goal is to find $\dot{\boldsymbol{z}}$ that maximally increases the loss (i.e. maximizes (\ref{eq:loss_pert})) under this penalty.
\begin{equation}
\dot{\boldsymbol{z}}^*_{\color{red} 3 N \times 1} \triangleq \arg\min_{\dot{\boldsymbol{z}}} \,-\, \langle \nabla_{\boldsymbol{z}} L(\boldsymbol{z}) \,,\, \dot{\boldsymbol{z}} \rangle \,+\, \frac{1}{2} \lambda \hat{P} \,.
\end{equation}
This leads to a a convex objective function\footnote{With abuse of notation, we have absorbed the constant $\delta_x \times \delta_y$ into $\lambda$.} in $\dot{\boldsymbol{z}}$ and its solution can be obtain by zero crossing the gradient,
\begin{eqnarray}
-\nabla_{\boldsymbol{z}_A} L(\boldsymbol{z}) + \lambda \boldsymbol{M}_A^T ( \boldsymbol{M}_A \dot{\boldsymbol{z}}_A + \boldsymbol{M}_B \dot{\boldsymbol{z}}_B  + \boldsymbol{M}_C \dot{\boldsymbol{z}}_C ) = \boldsymbol{0} \\
-\nabla_{\boldsymbol{z}_B} L(\boldsymbol{z}) + \lambda \boldsymbol{M}_B^T ( \boldsymbol{M}_A \dot{\boldsymbol{z}}_A + \boldsymbol{M}_B \dot{\boldsymbol{z}}_B  + \boldsymbol{M}_C \dot{\boldsymbol{z}}_C ) = \boldsymbol{0} \\
-\nabla_{\boldsymbol{z}_C} L(\boldsymbol{z}) + \lambda \boldsymbol{M}_C^T ( \boldsymbol{M}_A \dot{\boldsymbol{z}}_A + \boldsymbol{M}_B \dot{\boldsymbol{z}}_B  + \boldsymbol{M}_C \dot{\boldsymbol{z}}_C ) = \boldsymbol{0} \,.
\end{eqnarray}
The above form can be written more compactly as,
\begin{equation}
\boldsymbol{M}_{\color{red} 3N \times 3N} \dot{\boldsymbol{z}}_{\color{red} 3 N \times 1} = \frac{1}{\lambda} \nabla_{\boldsymbol{z}} L(\boldsymbol{z}) \,,
\end{equation}
where,
\begin{equation}
\boldsymbol{M}_{\color{red} 3N \times 3N} \triangleq \begin{pmatrix} \boldsymbol{M}_A^T \boldsymbol{M}_A & \boldsymbol{M}_A^T \boldsymbol{M}_B & \boldsymbol{M}_A^T \boldsymbol{M}_C \\
\boldsymbol{M}_B^T \boldsymbol{M}_A & \boldsymbol{M}_B^T \boldsymbol{M}_B & \boldsymbol{M}_B^T \boldsymbol{M}_C \\
\boldsymbol{M}_C^T \boldsymbol{M}_A & \boldsymbol{M}_C^T \boldsymbol{M}_B & \boldsymbol{M}_C^T \boldsymbol{M}_C
\end{pmatrix} \,.
\end{equation}
Thus,
\begin{equation}
\dot{\boldsymbol{z}}^*_{\color{red} 3N \times 1} = \frac{1}{\lambda} \boldsymbol{M}^{-1}_{\color{red} 3N \times 3N} \big(\nabla_{\boldsymbol{z}} L(\boldsymbol{z}) \big)_{\color{red} 3N \times 1} \,,
\end{equation}

While the matrix $\boldsymbol{M}$ is computed for each individual image, it is completely independent from the loss function $L$, and thus also independent of the model to be trained. This implies that for each image $\boldsymbol{z}$, the associated matrix $\boldsymbol{M}^{-1}$ can be computed once. Then, the computed $\boldsymbol{M}^{-1}$ can the be used for \emc{any learning model} in the future. Clearly $\boldsymbol{M}^{-1}$ captures some intrinsic properties of the data that are useful later for photometric transforms.

For the purposes of practical application, however, challenges still remain in its current formulation. Although the matrix $\boldsymbol{M}^{-1}$ associated with each image need only be computed once, and offline, the process of applying such a large matrix to the gradient vector can be expensive in terms of both computation and memory requirements. We thus use SVD to extract the leading eigenvectors of $\boldsymbol{M}^{-1}$ and use only those to achieve an efficient approximation. For example, Figure~\ref{fig:cow}-c shows the adversarial perturbation built from the top two leading eigenvectors of $\boldsymbol{M}^{-1}$. Equivalently, the eigenvectors corresponding to the smallest eigenvalues of $\boldsymbol{M}$ as an approximation. 

%% file: sections/experiments.tex
\begin{center}
\begin{table}[]
    \centering
    \begin{tabular}{ |c|c|c|c| } 
        \hline
         & CIFAR-10 & CIFAR-100 & STL-10\\
        \hline
        BasicAug & 95.56\% & 77.41\% & 84.60\% \\
        \hline
        AutoAug & 95.70\% & 77.41\% & 84.46\%\\
        \hline
        RandAug & 94.95\% & 74.56\% & 88.75\%\\
        \hline
        Flow + BasicAug & 95.79\% & 76.82\% & 85.41\%\\
        \hline
        Flow + AutoAug & 95.77\% & 77.26\% & 85.39\%\\
        \hline
        Flow + RandAug & 93.22\% & 70.14\% & 87.96\%\\
        \hline
        Recolor + BasicAug & 95.80\% & \textbf{77.45\%} & 86.11\%\\
        \hline
        Recolor + AutoAug & \textbf{95.86\%} & 76.97\% & 85.21\%\\
        \hline
        Recolor + RandAug & 94.78\% & 74.18\% & \textbf{88.84\%}\\
        \hline
    \end{tabular}
    \caption{Reported accuracy of a Wide-ResNet-32-10 model trained from scratch using the listed augmentations.}
    \label{tab:groundup}
\end{table}
\end{center}

\section{Experiments}
We compare the optical flow and recolorization augmentation schemes to two state-of-the-art data augmentation policies, AutoAugment and RandAugment, as well as a baseline with minimal augmentation.  We evaluate these methods on two types of experiments: ground-up training and finetuning.  Three datasets are considered, CIFAR-10, CIFAR-100, and STL-10 \cite{krizhevsky2009learning,coates2011analysis}. We use a validation set, with a 90-10 split, to select the choices of hyperparameters for the structured adversarial augmentations.  All final numbers were computed as the average of 10 runs.

\subsection{Ground-up Training}
In ground-up training, we train a Wide-ResNet-32-10 model \cite{zagoruyko2016wide} from scratch.  We use a learning rate of $10^{-1}$ with cosine decay, a batch size of 128, and perform training for 200 epochs.

\subsubsection{Augmentations} By default, images are processed using a base set of standard augmentation techniques: horizontal flips with 50\% probability, zero-padding, random crops, and 16x16 pixel cutout.  In our experiments, we apply the adversarial transformation with 50\% probability. The AutoAugment baseline utilizes a fixed, learned policy, and the RandAugment baseline is identical to the method described in the original paper \cite{cubuk2019randaugment}.

When considering stacking multiple augmentation strategies, ordering is significant.  In these experiments, the base augmentation is always applied last.  Adversarial transformations are applied first, directly on the raw image.  This modeling choice is made under the assumption that our method produces images that plausibly belong to the distribution over natural images.

\subsubsection{Results}  We report the test performance of models trained using structured adversarial perturbations combined with basic augmentation, combined with a parametric augmentation policy, as well as all baselines numbers.  The results are detailed in Table~\ref{tab:groundup}.  

The recolorization augmentation achieves the highest performance on all three datasets, albeit using different combinations.  Recolorization with AutoAugment achieves the best performance on CIFAR-10, beating the best baseline result, vanilla AutoAugment, by $0.16\%$.  It is noteworthy to point out that recolorization with basic augmentation, flow-transform with basic augmentation, and flow-transform with AutoAugment also outperform the baseline.  Recolorization with basic augmentation achieves the highest performance on CIFAR-100, beating the best baseline result by $0.04\%$.  Lastly, recolorization with RandAugment achieves the highest performance on STL-10, beating the best baseline result by $0.09\%$.  

Importantly, for each dataset, the baseline method that performed the best was improved upon with the addition of our augmentation scheme.  For CIFAR-10, the runner-up to recolorization with AutoAugment was vanilla AutoAugment.  This reinforces the idea that the addition of our augmentation strategy contributes to generalization.  Overall, these experimental results demonstrate that using structured adversarial augmentations such as recolorization and flow field transforms when training can successfully improve performance.

\subsection{Finetuning}

Powerful image classifiers have been demonstrated the ability to transfer performance to other image datasets with finetuning \cite{long2015fully,sharif2014cnn,huang2019gpipe,tan2019efficientnet}.  We thus evaluate the efficacy of our adversarial data augmentation techniques under this setting, and compare it to the baseline augmentation policies.

We take an EfficientNet-B0 checkpoint \cite{tan2019efficientnet}, and finetune it on the chosen dataset for 350 epochs.  In this experiment, no base augmentations are performed, and we do not search over combinations of augmentation strategies.  Also, whereas the AutoAugment and RandAugment policies are the same as in prior experiments, the structured adversarial augmentations are applied 100\% of the time rather than 50\% in their experiments.  This was empirically discovered to lead to higher performance.  The results of the experiments can be found in Table~\ref{tab:finetune}.

As shown, flow-based methods consistently outperform the baseline approaches across the CIFAR-10, CIFAR-100, and STL-10 datasets.  Applying the flow-based structured adversarial augmentations results in a $0.08\%$ improvement on CIFAR-10, a $0.19\%$ improvement on CIFAR-100, and a $0.06\%$ improvement on STL-10.  Finetuning with the recolorization augmentation slightly underperforms on the baselines, but matches the improved performance of applying flow-based augmentation on STL-10.

\begin{center}
    \begin{table}[]
        \centering
        \begin{tabular}{ |c|c|c|c| } 
            \hline
             & CIFAR-10 & CIFAR-100 & STL-10\\
            \hline
            No Augmentation & 97.52\% & 85.68\% & 97.55\% \\
            \hline
            AutoAugment & 97.44\% & 84.09\% & 97.32\%\\
            \hline
            RandAugment & 96.39\% & 80.38\% & 97.14\%\\
            \hline
            Flow & \textbf{97.60}\% & \textbf{85.87\%} & \textbf{97.61\%}\\
            \hline
            Recolorization & 97.34\% & 85.25\% & \textbf{97.61}\%\\
            \hline
        \end{tabular}
        \caption{Reported accuracy after finetuning on a EfficientNet-B0 checkpoint for 350 epochs.}
        \label{tab:finetune}
    \end{table}
\end{center}

\subsection{Targeted Adversarial Gradient}
In existing literature, construction of adversarial examples are categorized as either targeted or untargeted.  Targeted adversarial examples seek to transform the image such that the model increases its confidence in a desired label.  In the case of untargeted adversarial examples, the only requirement is that the model decreases its confidence in the correct label.

The adversarial gradient can be used to encode both untargeted and targeted augmentation information depending on how it is designed.  To create a targeted adversarial gradient, we simply take the gradient of the image with respect to a desired label.  Similarly, to create an untargeted adversarial gradient, we use the negative image gradient with respect to the correct label.  We explore and visualize the effect of using these two formulations in the case of a geometric transform on an image of a rugby ball.  A rugby ball is a natural fit for demonstrating the effect of geometric transform, as understanding its unique shape is crucial for the model to make a correct classification.

\begin{figure}
\includegraphics[width=1\linewidth]{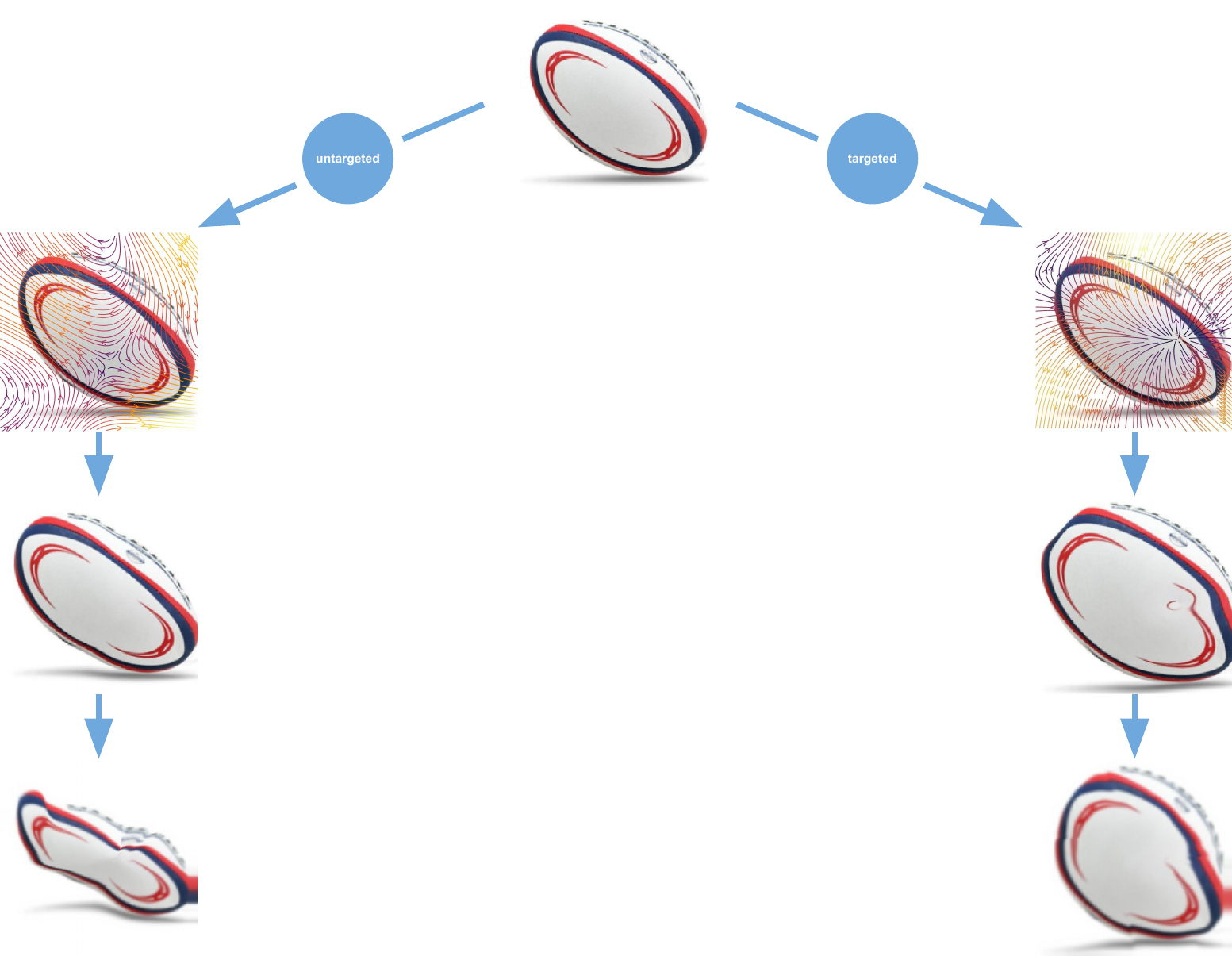}
\caption{Comparing targeted and untargeted flow transformations on a rugby ball.  The first layer shows the adversarial flow transformation overlaid on the original image.  The second layer shows the image generated by applying the flow.  The third layer shows the result of repeatedly generating and applying adversarial flows for ten iterations.}
\label{fig:rugby}
\end{figure}

At the top of Figure~\ref{fig:rugby}, we display the unmodified rugby ball. Our default model correctly classifies it with a confidence of 99.34\%.  On the left branch, we depict the adversarial flow transformation generated from the untargeted gradient.  We see that the arrows strongly push outwards along both ends of the major axis, and inwards along both ends of the minor axis.  Essentially this has the effect of deflating and elongating the rugby ball. The new image is still correctly classified by our model as a rugby ball, but the confidence has decreased to 97.94\%. To demonstrate the result from repeatedly applying our untargeted adversarial augmentation scheme, we arrive at the flattened ball in the third layer.  Now, the model believes the image to be a ski with 65.52\% confidence, and a rugby ball with 0.18\% confidence.  The shape of the ball has been flattened repeatedly, until it has been destroyed and can no longer be recognized.

On the other hand, the right branch displays the result of using a targeted adversarial gradient.  Instead of arbitrarily selecting a label to be our target, we select it in an adversarial way; we use the label out of all incorrect labels that the model places the most confidence in.  In this example, the model's most-confident incorrect label is a volleyball.  We see that the generated targeted adversarial flow pushes mass out within the ball, with the strongest directions along the minor axis.  Furthermore, at the top-left area of the ball, the flow field seeks to curb growth along the major axis.  This has the combined effect of deforming the rugby ball into a more spherical shape; in other words, closer to the structure of a volleyball.  The model's confidence in the rugby image has decreased to 96.69\%; however, the volleyball label confidence has increased from 0.30\% in the original image to 1.43\%.  When we repeatedly generate and apply these targeted transforms over ten iterations we find that halfway through, the next-best predicted label becomes a baseball.  The final result, shown in the bottom-most layer, depicts a very round object. The model places a confidence of 49.41\% in the baseball label, and a 26.43\% in the original rugby ball label.  Indeed, one can also visually verify the similarities the generated image shares with a baseball.

In practice however, we found no substantial performance improvement when using targeted over untargeted gradients.  As such, all experimental results in the paper are conducted using the untargeted scheme.

%% file: sections/discussion.tex
\section{Conclusion}

In this work, we propose structured adversarial perturbations for data augmentation. This was formulated as a constrained optimization task; while the objective seeks perturbations that maximally increase the loss, the constraint imposes specific structures on the perturbation to keep it closer to natural transformations.
 
One major limitation of this work is the cost of generating these operators.  In the case of recolorization, each image needs to generate its own unique operator.  This makes it difficult to scale to larger datasets.  Additionally, generating such operators on the fly is computationally expensive due to the large matrix inversion.  The flow transformation relies on only one operator, which makes it more tractable to extend to larger datasets.  However, the size of the operator is also a potential bottleneck; when handling large image dimensions, it can demand extremely large amounts of memory to store, and can take an enormous amount of time to compute.  Interesting future directions include learning these operators directly, or using approximations of these operators for our computations.